\documentclass{article}

\usepackage[preprint, nohyperref]{corl_2022} %

\usepackage[utf8]{inputenc}   %
\usepackage[T1]{fontenc}      %
\usepackage{booktabs}         %
\usepackage{amsfonts}         %
\usepackage{nicefrac}         %
\usepackage{microtype}        %
\usepackage[usenames,dvipsnames]{xcolor}           %
\usepackage{multirow}         %
\usepackage{verbatim}         %
\usepackage{ulem}             %
\usepackage{subcaption}
\usepackage{graphicx}
\usepackage{colortbl}         %
\usepackage[inline]{enumitem} %

\usepackage{xspace}           %
\usepackage{amsmath}
\usepackage[pagebackref=true,breaklinks=true,colorlinks=false]{hyperref}
\hypersetup{
    colorlinks,
    linkcolor={red!50!black},
    citecolor={blue!50!black},
    urlcolor=[RGB]{0,0.5,0.5}%
}

\title{Trajectory Forecasting on Temporal Graphs}

\author{
  G\"{o}rkay Aydemir$^1$\thanks{Work done during an internship at KUIS AI Center.} , \ Adil Kaan Akan$^2$, \ Fatma G\"{u}ney$^2$ \\
  $^1$ Middle East Technical University \quad
  $^2$ KUIS AI Center, Ko\c{c} University \\
  \texttt{gorkay.aydemir@metu.edu.tr}\quad\texttt{\{kakan20, fguney\}@ku.edu.tr}\\
  {\tt\small\textbf{\url{https://kuis-ai.github.io/ftgn}}}
  \vspace{-1cm}
}

\begin{document}

\newcommand{\Perp}{\perp\!\!\! \perp}
\newcommand{\bK}{\mathbf{K}}
\newcommand{\bX}{\mathbf{X}}
\newcommand{\bY}{\mathbf{Y}}
\newcommand{\bk}{\mathbf{k}}
\newcommand{\bx}{\mathbf{x}}
\newcommand{\by}{\mathbf{y}}
\newcommand{\bhy}{\hat{\mathbf{y}}}
\newcommand{\bty}{\tilde{\mathbf{y}}}
\newcommand{\bG}{\mathbf{G}}
\newcommand{\bI}{\mathbf{I}}
\newcommand{\bg}{\mathbf{g}}
\newcommand{\bS}{\mathbf{S}}
\newcommand{\bs}{\mathbf{s}}
\newcommand{\bM}{\mathbf{M}}
\newcommand{\bw}{\mathbf{w}}
\newcommand{\eye}{\mathbf{I}}
\newcommand{\bU}{\mathbf{U}}
\newcommand{\bV}{\mathbf{V}}
\newcommand{\bW}{\mathbf{W}}
\newcommand{\bn}{\mathbf{n}}
\newcommand{\bv}{\mathbf{v}}
\newcommand{\bwv}{\mathbf{wv}}
\newcommand{\bq}{\mathbf{q}}
\newcommand{\bR}{\mathbf{R}}
\newcommand{\bi}{\mathbf{i}}
\newcommand{\bj}{\mathbf{j}}
\newcommand{\bp}{\mathbf{p}}
\newcommand{\bt}{\mathbf{t}}
\newcommand{\bJ}{\mathbf{J}}
\newcommand{\bu}{\mathbf{u}}
\newcommand{\bB}{\mathbf{B}}
\newcommand{\bD}{\mathbf{D}}
\newcommand{\bz}{\mathbf{z}}
\newcommand{\bP}{\mathbf{P}}
\newcommand{\bC}{\mathbf{C}}
\newcommand{\bA}{\mathbf{A}}
\newcommand{\bZ}{\mathbf{Z}}
\newcommand{\bff}{\mathbf{f}}
\newcommand{\bF}{\mathbf{F}}
\newcommand{\bo}{\mathbf{o}}
\newcommand{\bO}{\mathbf{O}}
\newcommand{\bc}{\mathbf{c}}
\newcommand{\bm}{\mathbf{m}}
\newcommand{\bT}{\mathbf{T}}
\newcommand{\bQ}{\mathbf{Q}}
\newcommand{\bL}{\mathbf{L}}
\newcommand{\bl}{\mathbf{l}}
\newcommand{\ba}{\mathbf{a}}
\newcommand{\bE}{\mathbf{E}}
\newcommand{\bH}{\mathbf{H}}
\newcommand{\bd}{\mathbf{d}}
\newcommand{\br}{\mathbf{r}}
\newcommand{\be}{\mathbf{e}}
\newcommand{\bb}{\mathbf{b}}
\newcommand{\bh}{\mathbf{h}}
\newcommand{\bhh}{\hat{\mathbf{h}}}
\newcommand{\btheta}{\boldsymbol{\theta}}
\newcommand{\bTheta}{\boldsymbol{\Theta}}
\newcommand{\bpi}{\boldsymbol{\pi}}
\newcommand{\bphi}{\boldsymbol{\phi}}
\newcommand{\bpsi}{\boldsymbol{\psi}}
\newcommand{\bPhi}{\boldsymbol{\Phi}}
\newcommand{\bmu}{\boldsymbol{\mu}}
\newcommand{\bsigma}{\boldsymbol{\sigma}}
\newcommand{\bSigma}{\boldsymbol{\Sigma}}
\newcommand{\bGamma}{\boldsymbol{\Gamma}}
\newcommand{\bbeta}{\boldsymbol{\beta}}
\newcommand{\bomega}{\boldsymbol{\omega}}
\newcommand{\blambda}{\boldsymbol{\lambda}}
\newcommand{\bLambda}{\boldsymbol{\Lambda}}
\newcommand{\bkappa}{\boldsymbol{\kappa}}
\newcommand{\btau}{\boldsymbol{\tau}}
\newcommand{\balpha}{\boldsymbol{\alpha}}
\newcommand{\nR}{\mathbb{R}}
\newcommand{\nN}{\mathbb{N}}
\newcommand{\nL}{\mathbb{L}}
\newcommand{\cN}{\mathcal{N}}
\newcommand{\cM}{\mathcal{M}}
\newcommand{\cR}{\mathcal{R}}
\newcommand{\cB}{\mathcal{B}}
\newcommand{\cG}{\mathcal{G}}
\newcommand{\cL}{\mathcal{L}}
\newcommand{\cH}{\mathcal{H}}
\newcommand{\cS}{\mathcal{S}}
\newcommand{\cT}{\mathcal{T}}
\newcommand{\cO}{\mathcal{O}}
\newcommand{\cC}{\mathcal{C}}
\newcommand{\cP}{\mathcal{P}}
\newcommand{\cE}{\mathcal{E}}
\newcommand{\cI}{\mathcal{I}}
\newcommand{\cF}{\mathcal{F}}
\newcommand{\cK}{\mathcal{K}}
\newcommand{\cV}{\mathcal{V}}
\newcommand{\cY}{\mathcal{Y}}
\newcommand{\cX}{\mathcal{X}}
\def\bgamma{\boldsymbol\gamma}

\newcommand{\specialcell}[2][c]{%
  \begin{tabular}[#1]{@{}c@{}}#2\end{tabular}}

\newcommand{\figref}[1]{\Fig~\ref{#1}}
\newcommand{\secref}[1]{Section~\ref{#1}}
\newcommand{\algref}[1]{Algorithm~\ref{#1}}
\newcommand{\eqnref}[1]{Eq.~\eqref{#1}}
\newcommand{\tabref}[1]{Table~\ref{#1}}

\newcommand{\rulesep}{\unskip\ \vrule\ }

\newcommand{\KLD}[2]{D_{\mathrm{KL}} \Big(#1 \mid\mid #2 \Big)}

\renewcommand{\b}{\ensuremath{\mathbf}}

\def\mc{\mathcal}
\def\mb{\mathbf}

\newcommand{\T}{^{\raisemath{-1pt}{\mathsf{T}}}}

\makeatletter
\DeclareRobustCommand\onedot{\futurelet\@let@token\@onedot}
\def\@onedot{\ifx\@let@token.\else.\null\fi\xspace}
\def\eg{e.g\onedot} \def\Eg{E.g\onedot}
\def\ie{i.e\onedot} \def\Ie{I.e\onedot}
\def\cf{cf\onedot} \def\Cf{Cf\onedot}
\def\etc{etc\onedot} \def\vs{vs\onedot}
\def\wrt{wrt\onedot}
\def\dof{d.o.f\onedot}
\def\etal{et~al\onedot} \def\iid{i.i.d\onedot}
\def\Fig{Fig\onedot} \def\Eqn{Eqn\onedot} \def\Sec{Sec\onedot} \def\Alg{Alg\onedot}
\makeatother

\newcommand{\xdownarrow}[1]{%
  {\left\downarrow\vbox to #1{}\right.\kern-\nulldelimiterspace}
}

\newcommand{\xuparrow}[1]{%
  {\left\uparrow\vbox to #1{}\right.\kern-\nulldelimiterspace}
}

\renewcommand\UrlFont{\color{blue}\rmfamily}

\newcommand*\rot{\rotatebox{90}}
\newcommand{\boldparagraph}[1]{\vspace{0.2cm}\noindent{\bf #1:} }
\newcommand{\boldquestion}[1]{\vspace{0.2cm}\noindent{\bf #1} }

\newcommand{\ka}[1]{ \noindent {\color{blue} {\bf Kaan:} {#1}} } 
\newcommand{\ftm}[1]{ \noindent {\color{magenta} {\bf Fatma:} {#1}} }
\newcommand{\ga}[1]{ \noindent {\color{green} {\bf Gorkay:} {#1}} }

\maketitle

\begin{abstract}
Predicting future locations of agents in the scene is an important problem in self-driving. In recent years, there has been a significant progress in representing the scene and the agents in it. The interactions of agents with the scene and with each other are typically modeled with a Graph Neural Network. However, the graph structure is mostly static and fails to represent the temporal changes in highly dynamic scenes. In this work, we propose a temporal graph representation to better capture the dynamics in traffic scenes. We complement our representation with two types of memory modules; one focusing on the agent of interest and the other on the entire scene. This allows us to learn temporally-aware representations that can achieve good results even with simple regression of multiple futures. When combined with goal-conditioned prediction, we show better results that can reach the state-of-the-art performance on the Argoverse benchmark. 
\end{abstract}
\section{Introduction}
Self-driving is a complex task, therefore the standard approach is to divide it into separate modules. A typical modular pipeline consists of several modules focusing on different aspects of the problem such as perception, prediction, planning, and control. 
In this work, we assume that the perceptual input including the detections, the tracks, and the map information is provided in a bird's eye view representation, and focus on prediction by proposing a motion forecasting algorithm.
Motion forecasting is the problem of predicting the future location of traffic agents for safe navigation. This requires understanding the scene by representing the map information as well as the interactions of agents with the scene and with each other. Furthermore, there are multiple plausible future scenarios which need to be considered by the following planner module in the stack. %

Previous work on motion forecasting mostly focuses on learning representations of the scene, specifically the map and the agent history, \ie the previous locations of agents. While early attempts~\cite{Tang2019NeurIPS, Lee2017CVPR, Chai2019arXiv, Cui2019ICRA, Phan2020CVPR, Rhinehart2019CVPR, Casas2018CoRL, Hong2019CVPR, Biktairov2020NeurIPS} create a rasterized representation that can easily be processed with a 2D Convolutional Neural Network~(CNN), recent work mostly focuses on the spatial aspect with a lane graph~\cite{Liang2020ECCV, Gilles2021arXiv, Zeng2021IROS} or vector representations~\cite{Gao2020CVPR, Zhao2020arXiv, Gu2021ICCV}. An explicit representation of the topology and interactions with a Graph Neural Network~(GNN) leads to better representations of the surrounding environment as well as the agent interactions. In this work, we adapt the vectorized representation for the spatial aspect of the scene and then focus on the temporal aspect.

The temporal aspect is mostly ignored in motion forecasting by simply dividing the time into two: the present with the information up to now and the future to be predicted. Typically, the history of each agent is independently encoded with a recurrent neural network~\cite{Alahi2016CVPR, Gupta2018CVPR, Khandelwal2020arXiv, Mercat2020ICRA, Buhet2020CoRL, Park2020ECCV, Salzmann2020ECCV} or simply with a 1D CNN~\cite{Liang2020ECCV, Gilles2021ITSC, Gilles2021arXiv, Zeng2021IROS}. This approach fails to represent the evolving interactions between the agents and their relation with the scene elements through time. We claim that learning temporal dynamics plays a crucial role in prediction. Consider a scenario where two vehicles approach an intersection, their speed history changing together decides their future locations, for example one of them slowing down and letting the other vehicle continue with the turn at its current speed. Our results show improvements in these scenarios where the temporal dynamics are crucial for future prediction. 
Ye \etal~\cite{Ye2021CVPR} recently proposed to model the temporal aspect by focusing on the dynamics of the agent of interest only. While this improves the results, it overlooks the dynamics in the other parts of the scene that might still be relevant for predicting the next location of the agent of interest. 
We propose to learn a temporal graph representation that is aware of the entire scene dynamics.

We construct a temporal graph representing the dynamic scene with agents moving and interacting with the scene and with each other. We dynamically update the features of each scene element in a way informed by the other scene elements such as the other agents in the scene and nearby road segments. While these interactions are modelled with a static GNN in previous work, we model the interactions on the graph temporally by considering the time axis in the updates. In addition, we introduce two memory modules; one specific to the agent of interest and another to the entire scene. Our experiments show the importance of dynamic updates and the two types of memory modules.

Another aspect in motion forecasting is the multi-modality which can be addressed by predicting multiple futures. While there are various approaches that predict a heatmap~\cite{Gilles2021ITSC, Gilles2021arXiv, Gilles2021ICLR} or learn a distribution~\cite{Tang2019NeurIPS, Huang2020RAL, Lee2017CVPR, Yuan2019ICLR, Gupta2018CVPR, Deo2018IV} to sample from, we follow a simpler approach that is commonly used in the literature by generating a set of predictions and applying the loss only on the closest one during training.  
Common metrics used in motion forecasting evaluate both the quality and the diversity of endpoint predictions. As shown in recent work~\cite{Zhao2020arXiv, Gilles2021ITSC}, addressing these two aspects together is challenging. While one option is to develop separate objectives optimizing each metric, we instead focus on learning representations that are good at predicting accurate endpoint distributions without sacrificing diversity. 
\begin{figure}[t!]
\centering
\includegraphics[width=1\linewidth]{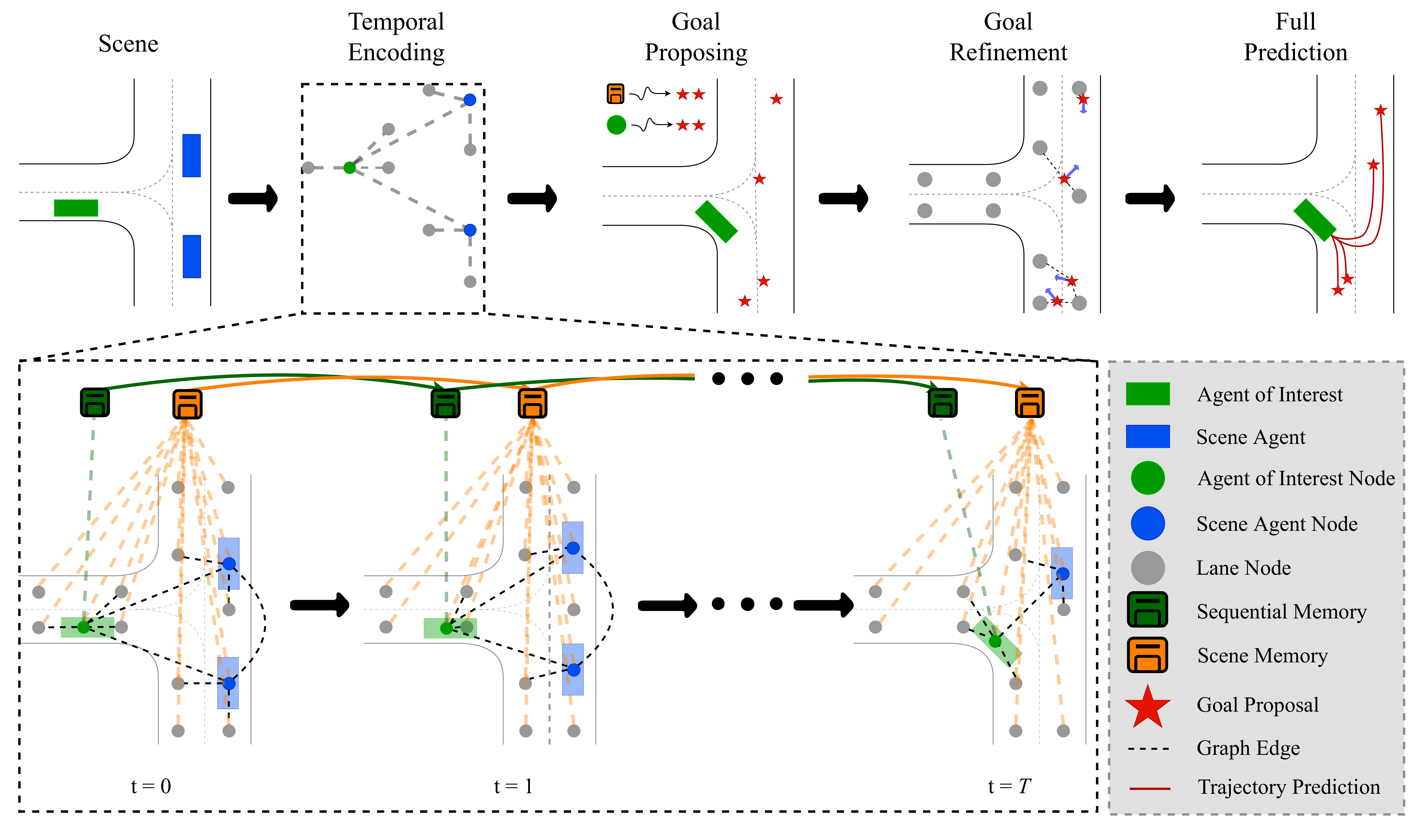}
\caption{\textbf{Temporal Graph Learning for Motion Forecasting.} We learn a dynamic scene representation where each timestamp is encoded as a temporal graph. We keep track of changes to the agent of interest with a sequential memory and to the entire scene with a scene memory. 
We generate goal proposals by using both the scene memory and the motion information related to the agent of interest.
Finally, we predict the full trajectories conditioned on the refined goal locations.}
\label{fig:overview} 
\vspace{-0.35cm}
\end{figure}

\section{Related Work}
\label{sec:rw}

\boldparagraph{Context Representation} 
Representing the context, \ie the surrounding environment is an important aspect of motion forecasting. Typically, the context consists of a map in 2D bird's eye view (BEV) representation as well as the past trajectories of agents on the map. There are two types of approaches to context representation: rasterized and vectorized.
In rasterized representation, the context is rastered and encoded with a 2D CNN. Despite the convenience of CNNs especially when predicting a heatmap~\cite{Gilles2021ITSC}, 2D raster image cannot explicitly represent the complex topology of the map such as long range connectivity and hierarchy between the scene elements due to the limited receptive field size. 
Vector representations initially proposed in VectorNet~\cite{Gao2020CVPR} can capture the complex topology of the road networks as well as the spatial locality of semantic entities with a hierarchical representation. Several followup work including ours~\cite{Zhao2020arXiv, Gu2021ICCV, Liu2021CVPR} build on VectorNet to represent the context. Differently, we also learn a temporal representation of the scene.

\boldparagraph{Temporal Encoding} Rossi~\etal propose to learn changes on a massive graph, \eg Wikipedia, Twitter, with a temporal graph neural network~\cite{rossi2020temporal}. We also propose to learn a temporal graph representation but for multiple, smaller graphs representing scenes observed through limited time intervals. The number of time steps are smaller but the changes are more dynamic through interactions between agents.
Recent work called TPCN~\cite{Ye2021CVPR} shows the importance of learning temporal relations in motion forecasting. Similar to us, in TPCN, there is a spatial module for a global representation and a temporal module for learning dynamics. Differently, multi-interval learning for temporal representation is specialized to the agent of interest in TPCN, whereas in our case, we learn temporal relations between all the entities. This way, learned dynamics can still help even when the interactions of the agent of interest is limited but there are other moving agents in the scene. Besides, predictions should be informed by the scene dynamics occurring in spatially further regions than the agent.

\boldparagraph{Goal-Conditioned Prediction} 
Earlier methods~\cite{Liang2020ECCV, Gao2020CVPR, Ye2021CVPR, Huang2021arXiv, Zeng2021IROS, Mercat2020ICRA, Song2021CoRL} directly predict $K$ full trajectories based on the features of the agent of interest. However, this approach may fail to cover diverse future locations on the map since it only focuses on the agent of interest. Some methods~\cite{Park2020ECCV} follow an auto-regressive approach which may lead to drift due to accumulating error in consecutive timestamps. Another line of work~\cite{Zhao2020arXiv, Liu2021CVPR, Gu2021ICCV, Gilles2021ITSC, Gilles2021ICLR, Gilles2021arXiv, Zhang2021arXiv} first predicts the endpoint of future trajectory and then conditioned on the predicted endpoint, the whole trajectory is predicted. We also follow this target-based approach because predicting trajectory is mostly straightforward once the target endpoint is identified.
The target-based methods~\cite{Zhao2020arXiv, Liu2021CVPR, Gu2021ICCV} typically follow a two-stage approach. First, a distribution over target locations is predicted, either to find the closest lane to the endpoint~\cite{Zhao2020arXiv} or densely over all possible locations on a grid~\cite{Gu2021ICCV}. Finding the closest lane is typically not accurate enough to locate the target point, therefore an offset is also predicted with respect to the lane~\cite{Zhao2020arXiv}. In this work, we show that our learned temporal representation is capable of directly regressing the target locations without scoring lanes or dense grid locations. Another option is to predict a heatmap representing the probability distribution of the target location~\cite{Gilles2021ITSC, Gilles2021ICLR, Gilles2021arXiv}. In heatmap-based approaches, the sampling strategy becomes very important. While it can be optimized for very low miss rates, it is difficult to optimize it together with the endpoint accuracy.

\section{Methodology}
Given the past states of agents in the scene and an HD map of the environment, our goal is to predict the future locations of the agent of interest. Our approach illustrated in \figref{fig:overview} is based on the following observation: traffic scenes consist of a dynamic part with agents moving through time and a static context which remains unchanged except for the interactions with the dynamic part. We first build a holistic representation of the static part and then model the dynamics as a temporal graph on top of it.

\subsection{Scene Encoding}
\subsubsection{Temporal Graph Representation}
\label{sec:temporal_graph_rep}
We construct a dynamic graph to represent the state of the scene at time $t$ as a graph $\cG_t = \{\cV_t, \cE_t \}$ where $\cV_t$ and $\cE_t$ denote the set of vertices and undirected edges on the graph. Each vertex $\bv^i_t \in \cV_t$ corresponds to a lane segment or an agent $i$ at time $t$. Due to the cost of a fully connected graph at each time step, we selectively build two types of edges between different types of nodes. We first connect each agent to the lane segments in their vicinity based on a threshold. The undirected edge between an agent and the surrounding lane segments allows to represent the current location of an agent on the map as well as the occupancy of map locations through time. We also connect the agent of interest to all the other agents at that timestamp to model dynamic interactions between agents.

Let $\bF_t$ denote the feature matrix at time $t$ where each row corresponds to a vertex $\bv^i_t$ and $d_k$ denote the length of key features. We perform dynamic updates on the features through time using self attention \cite{Vaswani2017NeurIPS} between the connected nodes:
\begin{align}
    \hat{\bF}_t = \mathrm{softmax}\left(\dfrac{\bF_{t-1}^Q \left(\bF_{t-1}^K\right)^T}{\sqrt{d_k}} \right) \bF_{t-1}^V
\end{align}
where $\bF_{t-1}^Q, \bF_{t-1}^K$ and $\bF_{t-1}^V$ are linear transformations of the feature matrix from the previous timestamp.
After initializing the node features from VectorNet, we accumulate temporal information to learn the dynamics.
We explicitly encode the time information to form the final feature matrix $\bF_t$:
\begin{align}
    \label{eq:temp_feat}
    \bff^i_t = g_1 \left(\hat{\bff}^i_t + \varphi_{\mathrm{time}}(t)\right)
\end{align}
where $\bff^i_t$ denotes the features of the agent $i$ at time $t$, $\varphi_{\mathrm{time}}(\cdot)$ is a time encoder as proposed in \cite{rossi2020temporal} and $g_1$ is simply a two-layer MLP. 

\subsubsection{Memory Modules}
\label{sec:memory_modules}
An important aspect of learning dynamics is building a memory to remember necessary information from past steps. In Temporal Graph Networks~\cite{rossi2020temporal}, Rossi \etal propose to keep track of changes with a memory for every node and edge on the graph. While it is shown to be crucial for node or edge addition or removal tasks in case of TGN, such a fine-grained memory module is not only infeasible in our case but also excessive for predicting the trajectory of a single agent of interest. On the other hand, to predict future reliably, the agent of interest needs to remember the changes in its representation as well as the changes to the whole scene through time. Therefore, we consider two types of memory modules in our temporal graph representation: one for the agent of interest and another for the whole scene.

Given the temporal features of the agent of interest $\bff_t$ at time $t$, we keep track of changes to its representation sequentially with a GRU: 
\begin{align}
    \label{eq:mem_agent}
    \bh_t^{\mathrm{seq}} = \mathrm{GRU}\left(\bff_t, \bh_{t-1}^{\mathrm{seq}}\right)
\end{align}
where $\bh_t$ refers the hidden state of the GRU. 
Note that we drop the superscript on the node features as we build the sequential memory model only for the agent of interest.

We introduce another memory module for the whole scene: scene memory.
We initialize the scene memory at time $t$ by applying a linear layer to the feature matrix at that time step: $\bM_t^{(0)} = g_0\left(\bF_t\right)$. We then build the scene memory module as layers of self attention operations \eqref{eq:scene_mem_layered_att} followed by layer normalization at each layer $l$ \eqref{eq:scene_mem_norm}. 
\begin{align}
    \label{eq:scene_mem_layered_att}
    \hat{\bM}_t^{(l)} &= \mathrm{softmax}\left( \dfrac{\bM_t^{(l-1), Q}\left(\bM_t^{(l-1), K}\right)^T}{\sqrt{d_k}}\right) \bM_t^{(l-1), V} \\
    \label{eq:scene_mem_norm}
    \bM_t^{(l)} &= \varphi_{\mathrm{norm}}\left(\hat{\bM}_t^{(l)}\right)
\end{align}

After the last layer $L$, we aggregate all the node features with a max pool operation \eqref{eq:scene_mem_pool} to summarize the relevant scene features in $\bm_t^{(L)}$ and then relate them across time with a GRU \eqref{eq:mem_scene}.
\begin{align}
    \label{eq:scene_mem_pool}
    \bm_t^{(L)} &= \varphi_{\mathrm{pool}}\left(\bM_t^{(L)}\right) \\ 
    \label{eq:mem_scene}
    \bh_t^{\mathrm{mem}} &= \mathrm{GRU}(\bm_t^{(L)}, \bh_{t-1}^{\mathrm{mem}})
\end{align}

In addition to the memory modules, we use cross attention between the temporally updated features of the agent of interest and the context elements including the lanes and the other agents following~\cite{Gu2021ICCV}. 
The final representation for the agent of interest contains the lane and agent features as the result of cross attention and the temporal features \eqref{eq:temp_feat} including their initialization for learning the change as well as the memory modules for the agent \eqref{eq:mem_agent} and the whole scene \eqref{eq:mem_scene}.

\subsection{Goal-Conditioned Trajectory Prediction}
\label{sec:goal_conditioned_pred}
Several previous work~\cite{Zhao2020arXiv, Liu2021CVPR, Gu2021ICCV} address multi-modality by predicting $K$ number of trajectories for a scene. The loss is calculated based on the prediction that has the closest endpoint to the ground truth endpoint.
A recent line of work ~\cite{Zhao2020arXiv, Liu2021CVPR} first predicts the endpoints as the target locations, and then, conditioned on the targets, predicts the full trajectory. 
We follow a similar goal-conditioned approach but in a more focused way on target locations. While the common approach in this line of work is to score a large number of map elements first, in some cases even densely~\cite{Gu2021ICCV}, and then refine them, it distributes the focus evenly to relevant target locations and irrelevant, distant locations on the map. Therefore, we first regress $K$ number of goal locations, and then focus on refinement and scoring.

\subsubsection{Goal Prediction}
\label{sec:goal_pred}
We initially predict $K$ number of goal locations. 
As indicated by the failure cases of the previous work~\cite{Ye2021CVPR}, goal locations need to be constrained by both the motion and the map information. We obtain the motion information from the features of the agent of interest as explained at the end of \secref{sec:memory_modules}. Although the agent of interest interacts with the scene, its features do not directly correspond to the scene elements. Therefore, for map constrained goal locations, we construct a map feature  $\bff^{\mathrm{map}}$ by max-pooling the feature of the lane nodes in the scene graph, $\bL$, and concatenating them with the updated scene memory from~\eqref{eq:mem_scene}: 
\begin{align}
    \label{eq:map_feature}
    \bff^{\mathrm{map}}_T &= g_2\left(
    \varphi_{\mathrm{pool}}\left(\bL_{T}\right),
    \varphi_{\mathrm{agg}}\left(\bh_{T}^{\mathrm{mem}} \right) \right)
\end{align}
where $T$ is the last observed time step before prediction and $g_2(\cdot)$ is a 2-layer MLP. We generate half of the proposals from the features of the agent of interest and the other half from the map features.

Once the proposals are created, we refine and score them in a way informed by the scene features. Our goal is to assign high scores to the proposals that are consistent with dynamic scene features. We first encode the proposals that are 2D coordinates and then apply cross attention with scene features $\bF_T$ to place it in our map representation before refinement and scoring. 
Our objective is to minimize the distance between the predicted goal location that is the closest to the ground truth with a smooth-$\mathnormal{L_1}$ loss and also to increase its score with respect to the other proposals with a cross entropy loss.

We predict the full trajectory conditioned on $K$ goal locations and apply a smooth-$\mathnormal{L_1}$ loss to minimize the distance between the ground truth and the predicted trajectory for the best goal prediction.

\section{Experiments}

\subsection{Experimental Setup}
\boldparagraph{Dataset}
We use Argoverse motion forecasting dataset~\cite{Chang2019CVPR} which has more than 300K sequences with the map information and the agent trajectories. Each sequence or a scene is 5 seconds long sampled at 10~Hz, corresponding to 50 time steps. 
Given the map information and the agent history in the first 2 seconds~(20 frames), the goal is to predict the trajectory of the agent of interest for the next 3 seconds,~(30 frames). We follow the original training, validation, and test splits.

\boldparagraph{Metrics}
We use four different metrics to evaluate our work. 
\begin{enumerate*}[label=(\roman*)]
\item \textbf{Minimum Average Displacement Error}~(min-ADE) is the average displacement error between the ground truth trajectories and the predicted trajectory that has the closest final step to the ground truth endpoint over all time steps.
\item \textbf{Minimum Final Displacement Error}~(min-FDE) is the displacement error between the best final step prediction and the ground truth final step.
$\mathrm{minADE}_K$ and $\mathrm{minFDE}_K$ refers to the minimum over $K$ predictions.
\item \textbf{Miss Rate}~(MR) is the percentage of the scenes where none of the predicted trajectory endpoints are not within a threshold ~(2 meters) to the ground truth endpoint.
\item \textbf{Brier-minFDE}~($\mathrm{b-minFDE}$) is the official metric of the challenge by also considering the probability distribution $p$ of a trajectory. Specifically, $\mathrm{b-minFDE}$ is calculated by adding a probability score, $(1-p)^2$ to the $\mathrm{minFDE}$ value. We report the values for $K=6$.
\end{enumerate*}

\boldparagraph{Training Details} We construct the graph by including the lanes that are closer than 50~meters to the agent of interest in terms of Manhattan distance. We normalize and rotate the scene with respect to the position and the orientation of the agent of interest. %
We create an edge between a lane segment and an agent if the distance between them is less than 2~meters. 
We train our models with a batch size of $64$ for 36 epochs. We use Adam optimizer~\cite{Kingma2015ICLR} with an initial learning rate of $1 \times 10^{-4}$ and divide it by 5 at the end of $24^{th}$ and $30^{th}$ epochs.
We randomly scale the scene by using a scale factor in the range of $[0.75, 1.25]$ and also apply random translations to the polylines for data augmentation. We initialize the global context encoder of our model from a pre-trained VectorNet.

\subsection{Ablation Study}
\begin{table}[t]
   \begin{center}
      \small
      \centering
      \def\arraystretch{1.2}
      \begin{tabular}{c c c | c c | c c c c}
         \hline
         \multicolumn{3}{c|}{Temporal Encoding} & \multicolumn{2}{c|}{Goal} & \multirow{2}{*}{ minADE$_{6}$} & \multirow{2}{*}{ minFDE$_{6}$} & \multirow{2}{*}{ MR$_{6}$} & \multirow{2}{*}{ $\text{b-minFDE}_6$} \\
         \cline{1-3} 
         \cline{4-5} TG & Seq & Scene & Goal Pred. & Goal Loss & & & \\
         \hline \hline
          & & & & & 0.81 & 1.32 & 0.15 & 1.94 \\
         \checkmark & & & & & 0.75 & 1.18 & 0.12 & 1.81 \\
         \checkmark & \checkmark & & & & 0.76 & 1.19 & 0.13 & 1.82 \\
         \checkmark & & \checkmark & & & 0.75 & 1.15 & 0.12 & 1.78 \\
         \checkmark & \checkmark & \checkmark & & & 0.74 & 1.14 & 0.12 & 1.77 \\
         \hline
         \checkmark & \checkmark & \checkmark & \checkmark & & 0.74 & 1.12 & 0.12 & 1.73 \\
         \checkmark & \checkmark & \checkmark & \checkmark & \checkmark & \textbf{0.73} & \textbf{1.08} & \textbf{0.10} & \textbf{1.68} \\
         \hline
      \end{tabular}
   \end{center}
   \caption{\textbf{Ablation Study.} We evaluate the contribution of each component proposed including the temporal graph~(\textbf{TG}), the sequential memory~({\textbf{Seq}}), the scene memory~(\textbf{Scene}), and the goal conditioning~({\textbf{Goal Pred.}}) with also an additional loss on the best endpoint prediction~({\textbf{Goal Loss}}).}
   \label{tab:component_ablation}
   \vspace{-0.5cm}
\end{table}
\boldquestion{What is the contribution of each component proposed?}
We perform an ablation study to measure the effect of each component on the performance in \tabref{tab:component_ablation}. Adding the temporal graph (\textbf{TG}; \secref{sec:temporal_graph_rep}) significantly improves the performance in each metric compared to the VectorNet initialization in the first row. This shows the importance of learning temporal dynamics in the scene. 
On top of the temporal graph, we measure the effect of two types of memory modules~(\secref{sec:memory_modules}). The scene memory provides temporal information about the overall scene and the sequential memory about temporal dynamics related to the agent of interest such as speed changes. The sequential memory for the agent of interest only~(\textbf{Seq}) degrades the performance slightly but the scene memory~(\textbf{Scene}) improves it, and using them together results in the best performance. This shows the importance of propagating information from past time steps together with scene information.

\begin{table*}[h!]
    \begin{center}
    \begin{tabular}{l||c  c  c  c}
      & $\text{minADE}_6$ & $\text{minFDE}_6$ & $\text{MR}_6$ \\
      \hline
      \hline
      TNT \cite{Zhao2020arXiv} & 0.88 & 1.63 & 0.22 \\
      DenseTNT \cite{Gu2021ICCV} & 0.78 & 1.25 & 0.13 \\
      HOME \cite{Gilles2021ITSC} & - & 1.26 & 0.13 \\
      TPCN \cite{Ye2021CVPR} & \textbf{0.73} & 1.15 & \textbf{0.11} \\
      \hline
      Ours & \underline{0.74} & \textbf{1.14} & \underline{0.12} \\
      \hline
    \end{tabular}
    \end{center}
    \caption{\textbf{Performance based on Feature Representation.} We compare the performance of different methods by simply regressing multiple outputs without any target sampling or goal conditioning to highlight the importance of learned representations. Temporal representation learning methods, \ie TPCN and ours, outperform the others, showing the importance of learning temporal dynamics.}
    \label{tab:regression_result}
    \vspace{-0.5cm}
\end{table*}
\boldquestion{How useful is the learned temporal representation?}
In order to measure the effect of learning a temporal representation, in \tabref{tab:regression_result}, we compare our method to the other methods by discarding the effect of goal conditioning in target-based methods~\cite{Zhao2020arXiv, Gu2021ICCV} and target sampling in a heatmap-based method~\cite{Gilles2021ITSC}. Following the previous work on representation learning~\cite{chen2020ICML, grill2020neurips} where an MLP is trained on top of a frozen backbone to measure the quality of a learned representation, we train an MLP on top of our backbone to directly regress $K$ trajectories. The more informative the features extracted by the backbones are, the better the predicted trajectories will be. As can be seen from \tabref{tab:regression_result}, the two methods, TPCN~\cite{Ye2021CVPR} and ours, which learn temporal representations outperform the other methods. This shows the importance of learning dynamics independent of other factors.

\boldquestion{What is the role of goal prediction?}
In the upper part of \tabref{tab:component_ablation}, we basically regress $K$ trajectories directly. In the bottom part, we measure the importance of goal prediction by first predicting $K$ goal locations and then predicting the full trajectories conditioned them. Goal conditioning improves the performance in terms of both minFDE and b-minFDE. Predicting the endpoint accurately is crucial since we condition on it to predict the trajectory next. Therefore, we apply an additional goal loss on the best endpoint directly which results in the best performance in all metrics. 
We also ablate the source, \ie which features to use, for goal prediction (Supplementary) and find that using both the map and the motion information improves the results in terms of MR and b-minFDE.

\subsection{Comparison to Previous Work}
\begin{table*}[t]
    \begin{center}
    \begin{tabular}{l||c  c  c >{\columncolor[gray]{0.8}}c}

      & $\text{minADE}_6$ & $\text{minFDE}_6$ & $\text{MR}_6$ & $\text{b-minFDE}_6$ \\
      \hline
      \hline
      HOME \cite{Gilles2021ITSC} & 0.94 & 1.45 & 0.10 & - \\
      Autobot \cite{Girgis2021arXiv} & 0.89 & 1.41 & 0.16 & - \\
      TPCN \cite{Ye2021CVPR} & 0.87 & 1.38 & 0.16 & - \\
      LaneRCNN \cite{Zeng2021IROS} & 0.90 & 1.45 & \underline{0.12} & 2.15 \\
      Jean \cite{Mercat2020ICRA} & 1.00 & 1.41 & 0.13 & 2.12 \\
      PRIME \cite{Song2021CoRL} & 1.22 & 1.56 & \underline{0.12} & 2.10 \\
      LaneGCN \cite{Liang2020ECCV} & 0.87 & 1.36 & 0.16 & 2.05 \\
      mmTransformer \cite{Liu2021CVPR} & \underline{0.84} & 1.34 & 0.15 & 2.03\\
      DenseTNT \cite{Gu2021ICCV} & 0.88 & \underline{1.28} & 0.13 & 1.98 \\
      THOMAS \cite{Gilles2021ICLR} & 0.94 & 1.44 & \textbf{0.10} & 1.97 \\
      Scene Transformer \cite{Ngiam2021ICLR} & \textbf{0.80} & \textbf{1.23} & 0.13 & \textbf{1.89} \\
      \hline
      Ours & 0.86 & 1.31 & 0.15 & \underline{1.93} \\
      \hline
    \end{tabular}
    \end{center}
    \caption{\textbf{Results on Argoverse Leaderboard (Test Set).} Our method is among the top-performing methods, the second in main ranking metric b-minFDE$_6$, and top-3 in minADE$_6$, minFDE$_6$.}
    \label{tab:argo_test}
    \vspace{-0.3cm}
\end{table*}

We compare our method's performance with the other published methods on both the validation set~(\tabref{tab:argo_val}) and the leaderboard~(\tabref{tab:argo_test}). Our method is among the top-performing methods, top-3 in minADE and minFDE on both the validation and test, which shows the endpoint prediction and trajectory completion accuracy, and the second best in the official ranking metric b-minFDE. Note that our method can obtain competitive performance in all metrics without optimizing for any metric specifically. Without targeting the MR specifically, our method can reach $10\%$ on the validation set which is the third best. This is because our method can predict a good endpoint distribution as shown by the second best b-minFDE on the test set.

\begin{table*}[h!]
    \vspace{-0.2cm}
    \begin{center}
    \begin{tabular}{l||c  c  c  c}
      & $\text{minADE}_6$ & $\text{minFDE}_6$ & $\text{MR}_6$ \\
      \hline
      \hline
      HOME \cite{Gilles2021ITSC} & - & 1.28 & 0.07 \\
      TPCN \cite{Ye2021CVPR} & 0.73 & 1.15 & 0.11 \\
      LaneRCNN \cite{Zeng2021IROS} & 0.77 & 1.19 & \underline{0.08} \\
      PRIME \cite{Song2021CoRL} & - & - & \underline{0.08} \\
      LaneGCN \cite{Liang2020ECCV} & \textbf{0.71} & \underline{1.08} & - \\
      mmTransformer \cite{Liu2021CVPR} & \textbf{0.71} & 1.15 & 0.11 \\
      DenseTNT-MR$^\dagger$ \cite{Gu2021ICCV} & {0.82} & {1.28} & \textbf{0.07} \\
      DenseTNT-minFDE$^\dagger$ \cite{Gu2021ICCV} & {0.76} & \textbf{1.05} & {0.10} \\
      \hline
      Ours & \underline{0.73} & \underline{1.08} & 0.10 \\
      \hline
    \end{tabular}
    \end{center}
    \caption{\textbf{Results on Argoverse Validation Set.} Our method is among the top-performing methods, the second in minADE$_6$, minFDE$_6$, top-3 in MR$_6$. $^\dagger$reproduced results using the official implementation.}
    \label{tab:argo_val}
    \vspace{-0.3cm}
\end{table*}

\subsection{Qualitative Results}
\vspace{-0.2cm}
\begin{figure}[t!]
\centering
\begin{subfigure}[b]{0.3\textwidth}
   \includegraphics[width=\linewidth]{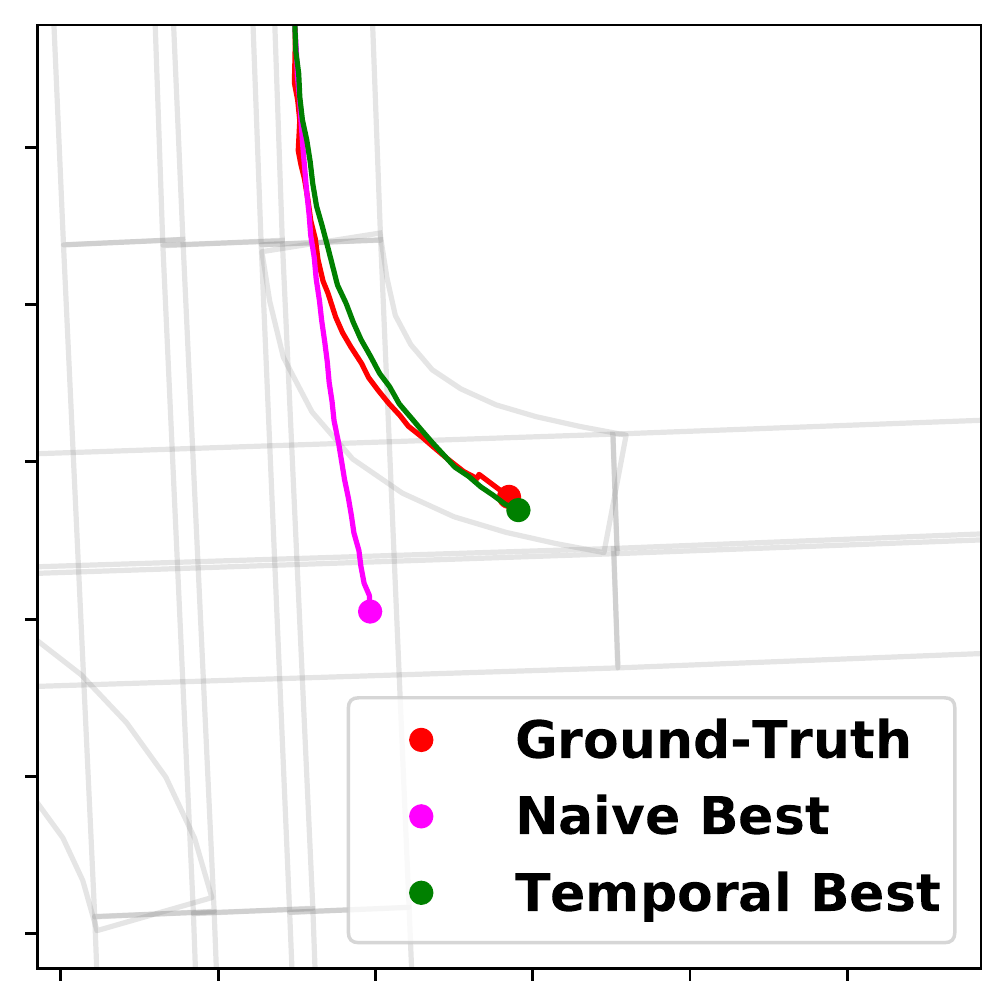}
   \caption{}
   \label{fig:map_comp1} 
\end{subfigure}
\hfill
\begin{subfigure}[b]{0.3\textwidth}
    \includegraphics[width=\linewidth]{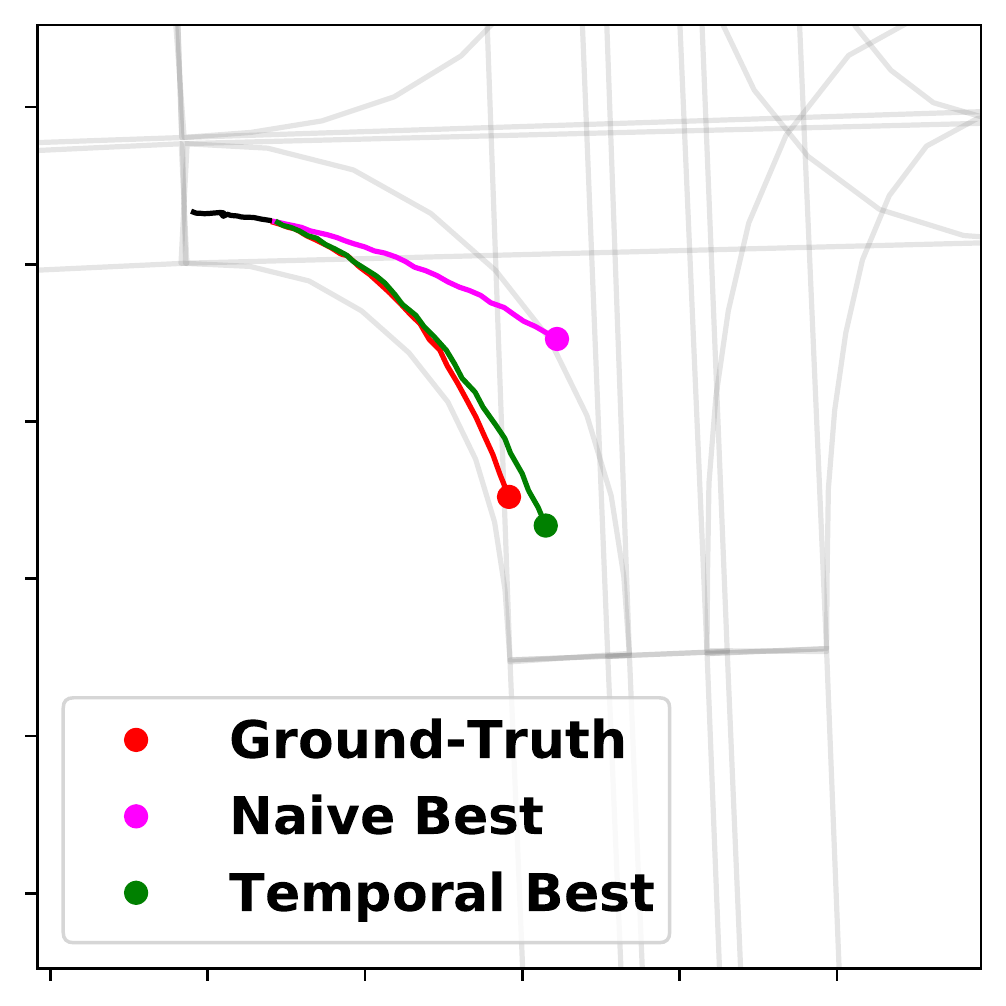}
    \caption{}
    \label{fig:map_comp2} 
\end{subfigure}
\hfill
\begin{subfigure}[b]{0.3\textwidth}
    \includegraphics[width=\linewidth]{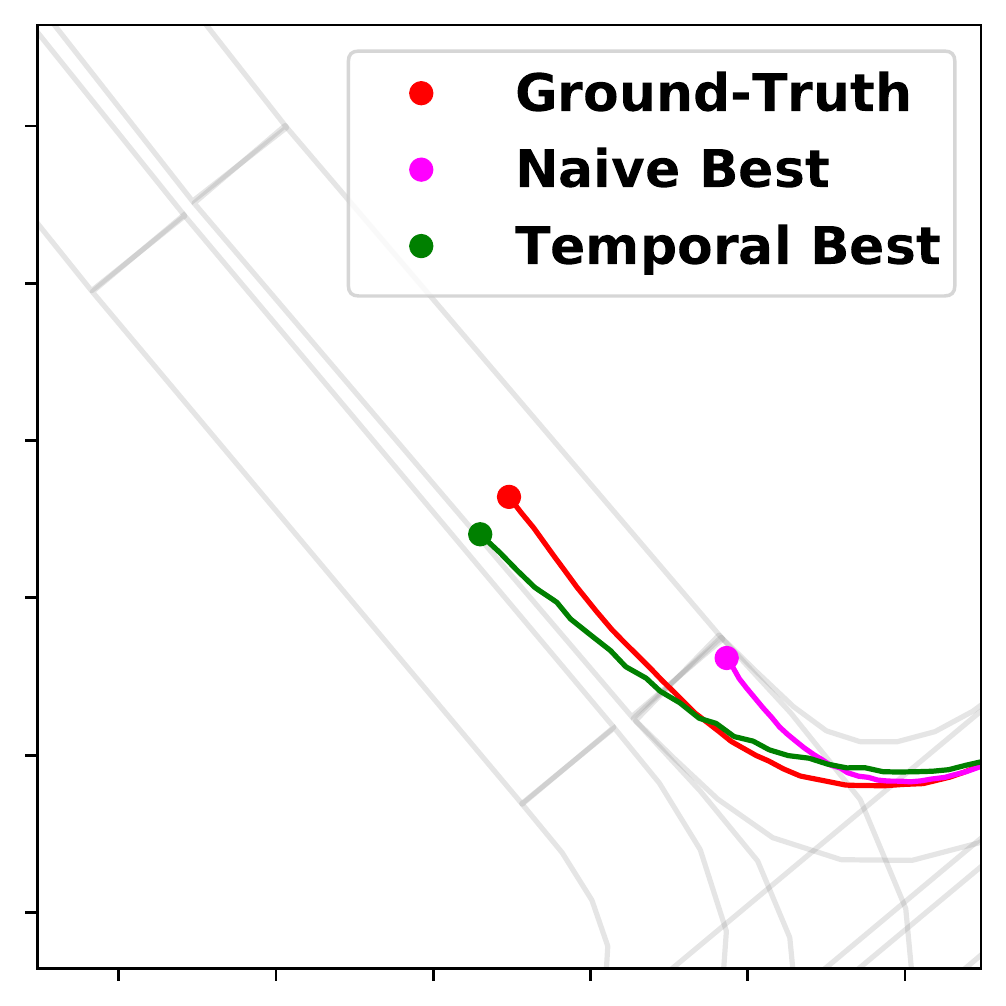}
    \caption{}
    \label{fig:speed_comp1}
\end{subfigure}
\\
\begin{subfigure}[b]{\textwidth}
  \includegraphics[width=1\linewidth]{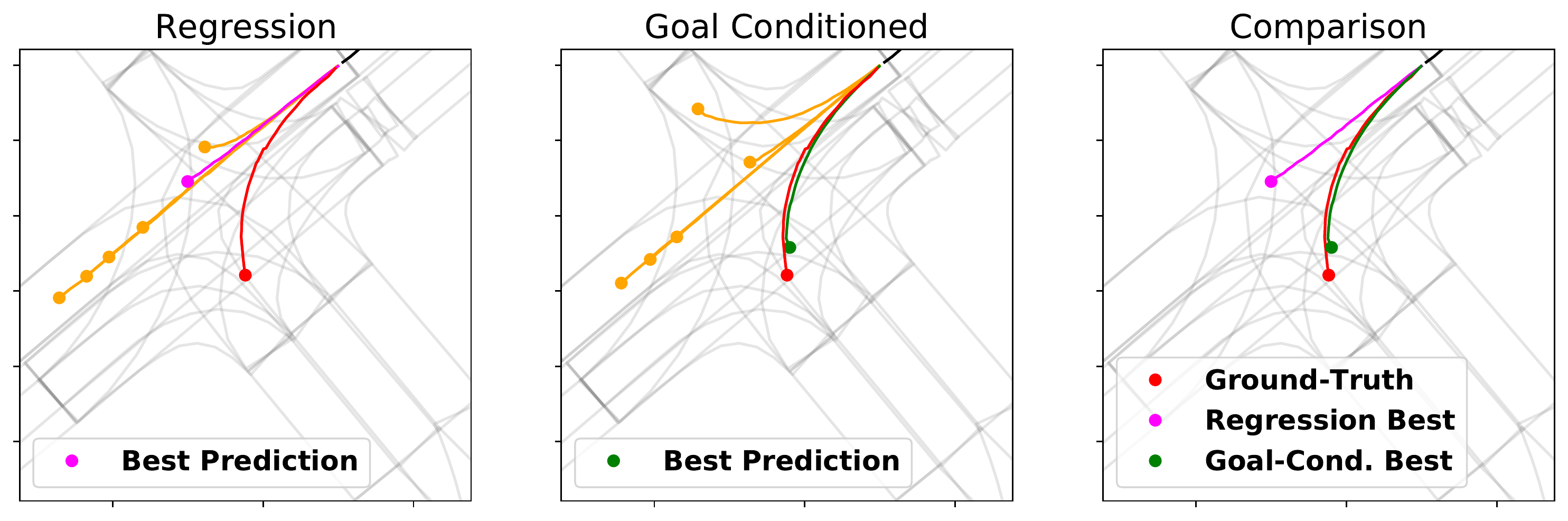}
  \caption{}
  \label{fig:mode_comp1}
\end{subfigure}

\caption{\textbf{Temporal Graph Representation with Goal Conditioning.} We visualize the effect of learning temporal dynamics with our model~(\textbf{Temporal}) to a basic version without temporal graph~(\textbf{Naive}). Learning temporal dynamics leads to more admissible predictions agreeing with the map~(\subref{fig:map_comp1}, \subref{fig:map_comp2}) and better capturing the changes to the velocity~(\subref{fig:speed_comp1}). We also show that goal conditioning results in better distributions by covering more modes compared to simple regression (\subref{fig:mode_comp1}).}
\label{fig:qual_temporal}
\vspace{-0.5cm}
\end{figure}
We visualize the effectiveness of our temporal graph representation in \figref{fig:qual_temporal} where we compare our model~(\textbf{Temporal}) to a basic version without temporal graph~(\textbf{Naive}) which is similar to vanilla VectorNet~\cite{Gao2020CVPR} for global context encoding. Learning a temporal representation allows our model to generate more admissible trajectories respecting the borders of the map as shown in (\ref{fig:map_comp1}) and (\ref{fig:map_comp2}) and also to better capture the changes in speed as shown in (\ref{fig:speed_comp1}). In general~(see Supplementary for more examples), the improvements are more pronounced at the intersections where the map information is crucial and there are typically significant alterations to the speed of the agents.

In (\ref{fig:mode_comp1}), we compare our goal conditioned prediction to simple regression without any goal conditioning by visualizing all the $K$ trajectories predicted in orange. Even though simple regression performs well quantitatively~(\tabref{tab:component_ablation}), we can cover more modes with goal conditioning. %
The simple regression misses the left turn. With goal conditioning, we can not only predict the left turn but also the possibility of a right turn as well as going straight, all in agreement with the map. %

\section{Conclusion, Limitations, and Future Work}
We propose a temporal graph representation for motion forecasting  with two types of memory modules. Our method is among the top-performing methods on Argoverse, especially in terms of the official metric b-minFDE which measures the quality of the distributions.
We address diversity with a simple goal conditioning, therefore our method is not among the top-performing methods in terms of MR. In future, we plan to focus on diversity by extending our temporal graph to a probabilistic formulation.
Contemporary work~\cite{Ngiam2021ICLR} shows the importance of learning a holistic representation for the scene rather than focusing on a single agent. In this work, we still focus on a single agent in prediction but we consider temporal relations for the whole scene. An interesting future work can build on our temporal representation to improve predictions for all the agents in the scene.
In motion forecasting, algorithms rely on map info and perception input which may not be available in real life.
\bibliography{bibliography_long, egbib}
\clearpage
\appendix

In this material which is the supplementary of the paper Trajectory Forecasting on Temporal Graphs, we explain our implementation and training settings in detail, provide quantitative results related to goal conditioning mentioned in the paper and comprehensive qualitative results including component comparison and fail cases with reasons.
\section{Implementation Details}
In this section, we provide the details of our model for reproducibility. We will also publish our code when this work is published.

\textbf{Scene Representation:} In scene representation, we converted lanes and agent trajectories in vector form as in original VectorNet paper ~\cite{Gao2020CVPR}. We included all lanes that are closer than 50 meters to any agent in any past timestamp to make lane-agent interaction possible during temporal encoding. We kept the the feature vector size as 128. 

\textbf{Encoders:} Both MLP inputs and outputs are vectors of size 128 except enhanced agent of interest feature. As explained in the paper, we used cross attention between all nodes and and lane nodes and concatenated them form the final enhanced agent of interest feature of size 384. We extracted features of 2D points to vectors of size 128 with point subgraph of DenseTNT ~\cite{Gu2021ICCV} which is 3 linear layers taking inputs of a 2D point and agent feature. We followed cross attention mechanism of DenseTNT ~\cite{Gu2021ICCV} to get final point features.

\textbf{Graph Networks:} We set layer number to 3 in subgraphs of VectorNet backbone and Scene Memory Encoder GNN where each layer is 2 layer MLP. We $L_2$ normalized Subgraph outputs and did not use map completion loss. Global graphs of VectorNet backbone and Temporal Graph is implemented as a single head self attention. If there is no edge between nodes in temporal graph, we set probability of node inclusion to 0 by using adjacency matrix of a timestamp as the mask before softmax operation as proposed in original transformer \cite{Vaswani2017NeurIPS}. 

\textbf{Goal Conditioned Prediction:} We conditionally predicted the full trajectory with 2 layer MLP whose input is concatenation of the agent of interest feature and endpoint feature. To Similarly, we predicted endpoints by giving point features to 2 layer MLP.

\textbf{Data Augmentations:} Since we normalized the scene according to agent of interest, to scale the scene, we just multiplied the coordinate points with a value between $[0.75, 1.25]$. We added noise sampled from $\mathcal{N}(0, 0.2)$ to polyline locations as perturbation. 

\textbf{Training Details:} As mentioned in the paper, we trained our models with a batch size of $64$ for 36 epochs  corresponding to $115848$ iterations in 4 Tesla T4 GPUs in a distributed manner. We used Adam optimizer~\cite{Kingma2015ICLR} with an initial learning rate of $1 \times 10^{-4}$ and divide it by 5 at the end of $24^{th}$ and $30^{th}$ epochs.

\section{Quantitative Results}
\textbf{Does it matter which features we use for goal prediction?}
\begin{table}[ht!]
   \begin{center}
      \small
      \centering
      \def\arraystretch{1.2}
      
      \begin{tabular}{ c c | c c c c}
        \hline
        \multicolumn{2}{c|}{Target Source} & \multirow{2}{*}{ minADE$_{6}$} & \multirow{2}{*}{ minFDE$_{6}$} & \multirow{2}{*}{ MR$_{6}$} & \multirow{2}{*}{ $\text{b-minFDE}_6$} \\
        \cline{1-1} 
        \cline{2-2} Context & Agent & & &  \\
        \hline \hline
        \checkmark & & 0.73 & 1.08 & 0.11 & 1.69 \\
         & \checkmark & 0.73 & 1.09 & 0.11 & 1.69 \\
        \checkmark & \checkmark & 0.73 & 1.08 & 0.10 & 1.68 \\

        \hline
      \end{tabular}
      
   \end{center}
   \caption{\textbf{Comparing Features for Goal Prediction}. We compare the sources for goal prediction, including the map information~(\textbf{Context}), the motion information of the agent of interest~(\textbf{Agent}), and both where the half of the proposals come from the context and the other half from the agent.}
   \label{tab:target_source}
   \vspace{-5px}
\end{table} 
As explained in the methodology of the paper, we propose to use both the map information and the motion information about the agent of interest for predicting goal locations. In \tabref{tab:target_source}, we ablate this decision by comparing the performance using the map information only~(\textbf{Context}), the motion information only~(\textbf{Agent}), and using both as proposed. As can be seen from \tabref{tab:target_source}, using both improves the results in terms of MR and b-minFDE.
\section{Qualitative Results}
\subsection{Component Comparison}
\begin{figure}[h]
\centering
    \vspace{1cm}
    \begin{subfigure}[b]{\textwidth}
      \includegraphics[width=1\linewidth]{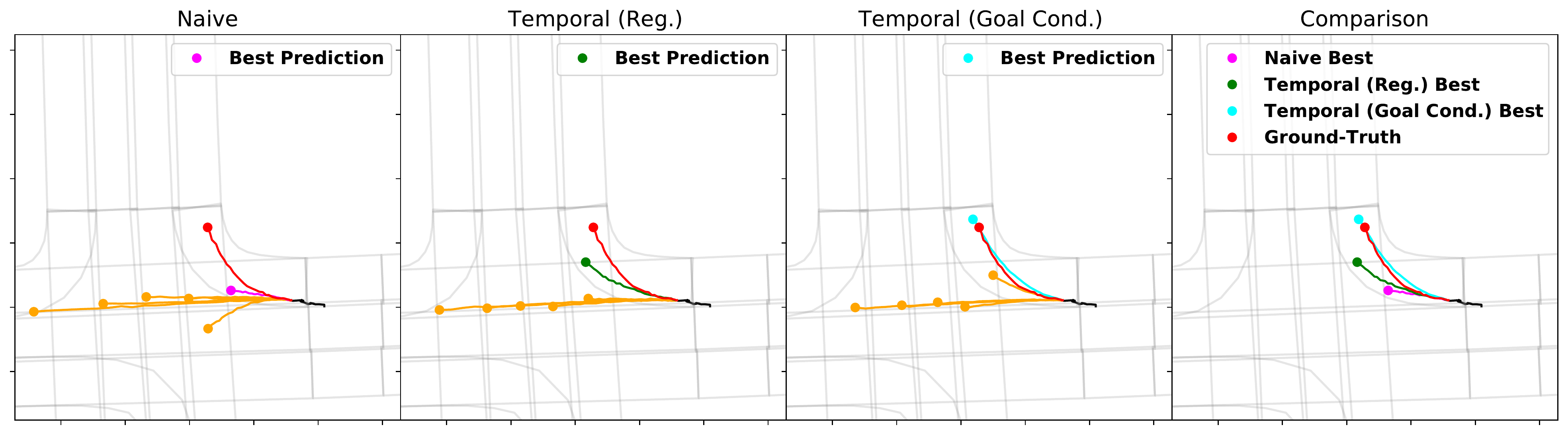}
    \end{subfigure}
    
    \begin{subfigure}[b]{\textwidth}
      \includegraphics[width=1\linewidth]{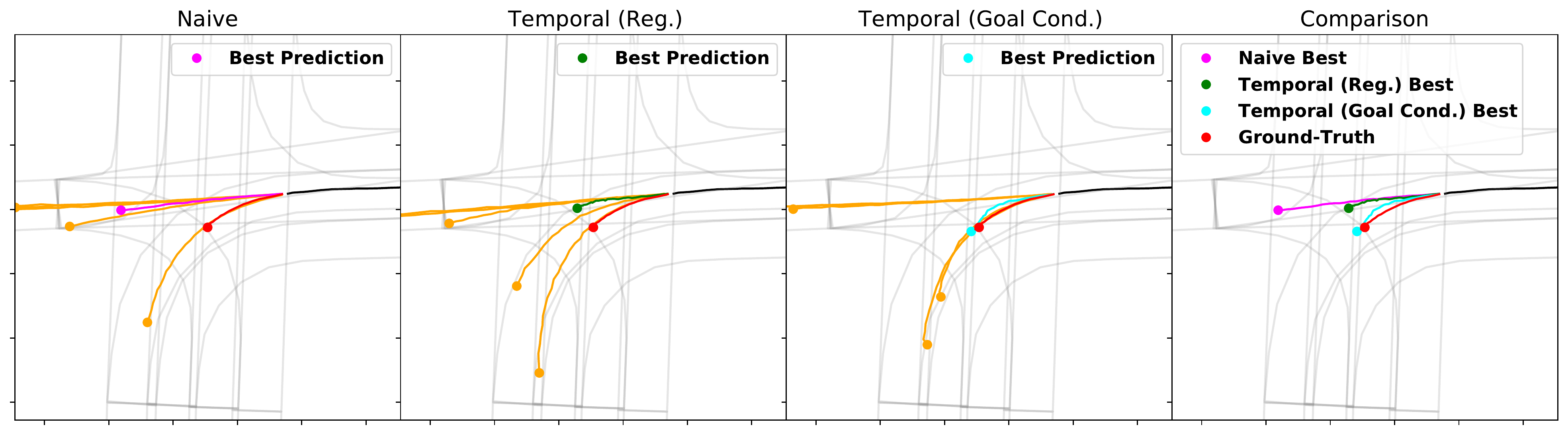}
    \end{subfigure}
    
\caption{\textbf{Component Comparison (a)}}
\label{fig:supp_comp_set1}
\end{figure}

\begin{figure}[h!]
\centering
    \begin{subfigure}[h]{\textwidth}
      \includegraphics[width=1\linewidth]{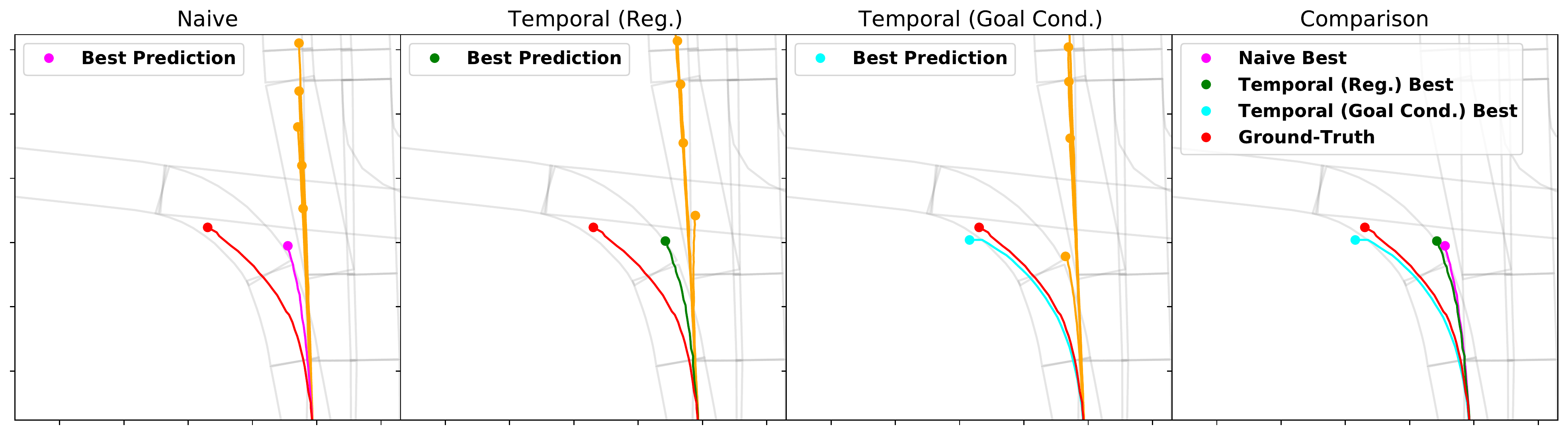}
    \end{subfigure}

    \begin{subfigure}[h]{\textwidth}
      \includegraphics[width=1\linewidth]{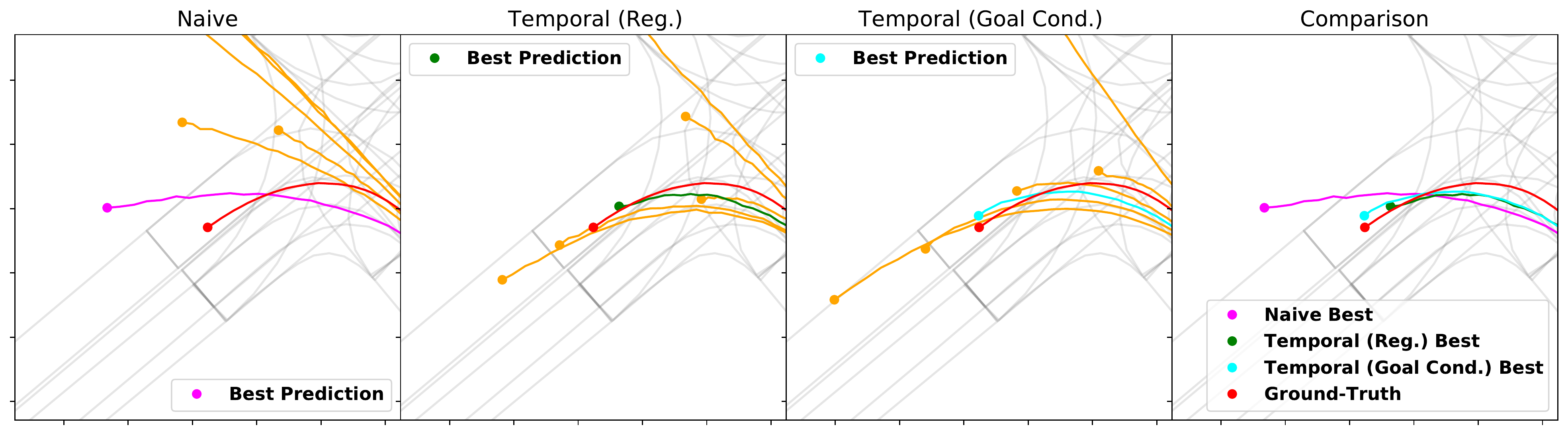}
    \end{subfigure} 
    
    \begin{subfigure}[h]{\textwidth}
      \includegraphics[width=1\linewidth]{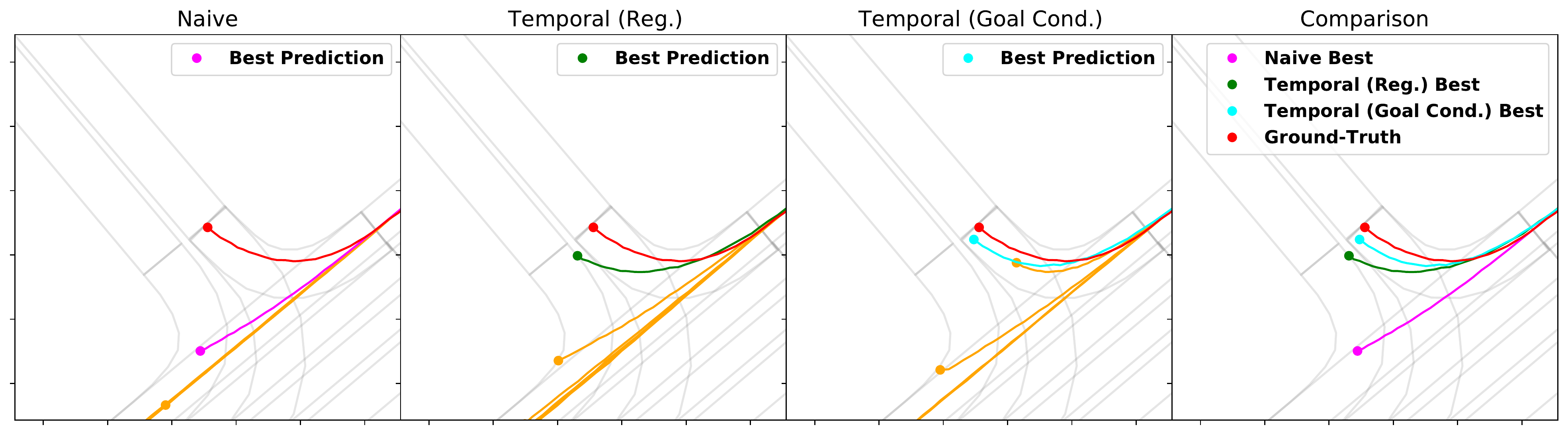}
    \end{subfigure} 

    \begin{subfigure}[h]{\textwidth}
      \includegraphics[width=1\linewidth]{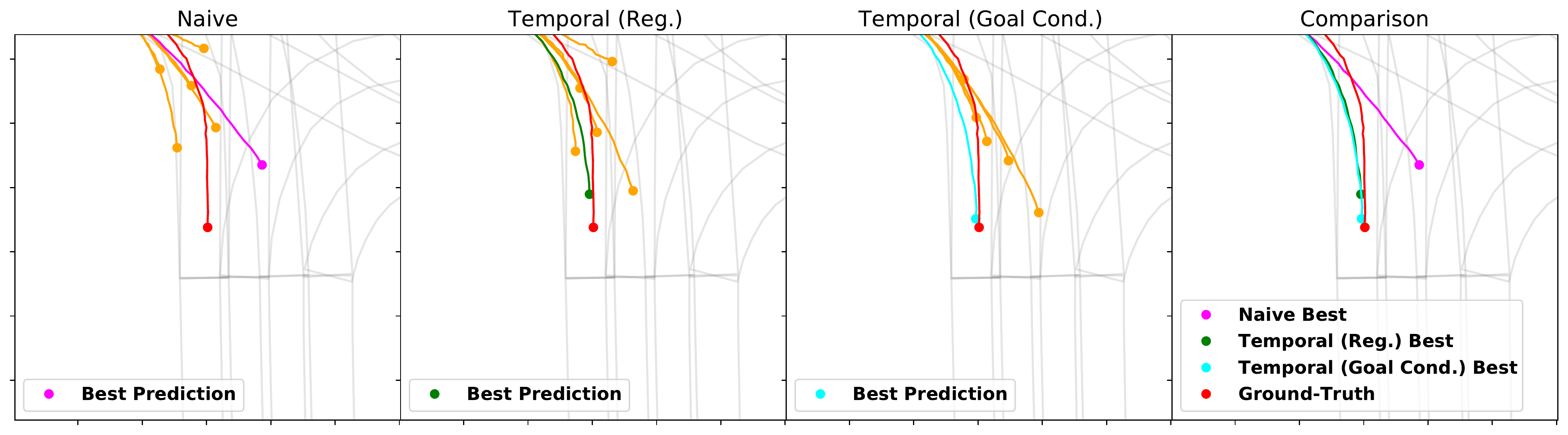}
    \end{subfigure} 

    \begin{subfigure}[h]{\textwidth}
      \includegraphics[width=1\linewidth]{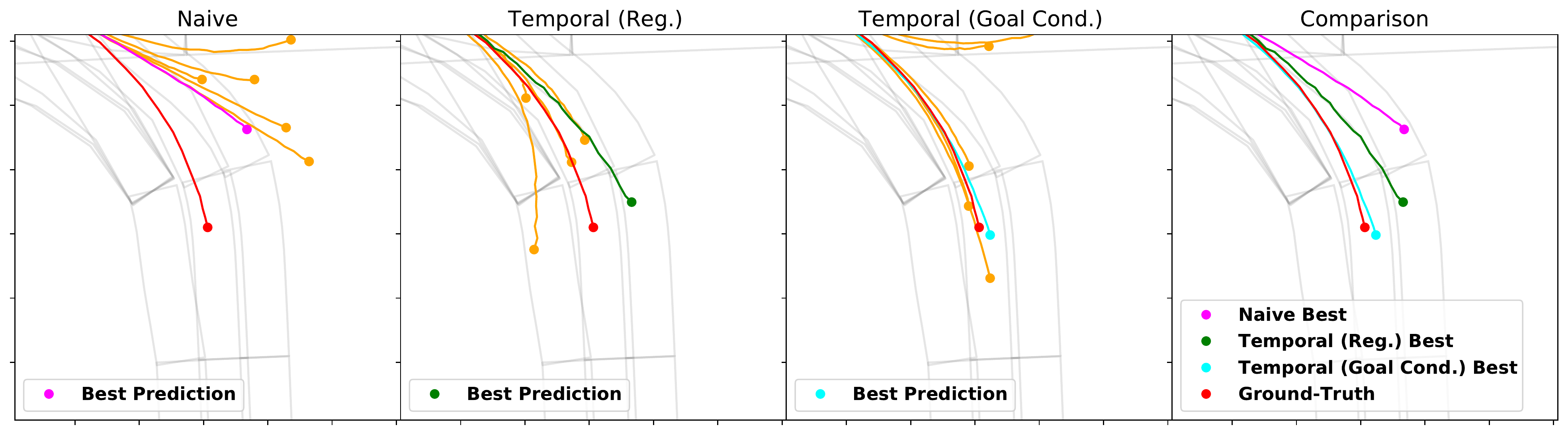}
    \end{subfigure} 
    
\caption{\textbf{Component Comparison (b)}}
\label{fig:supp_comp_set2}
\end{figure}

\begin{figure}[h!]
\centering
    \begin{subfigure}[h]{\textwidth}
      \includegraphics[width=1\linewidth]{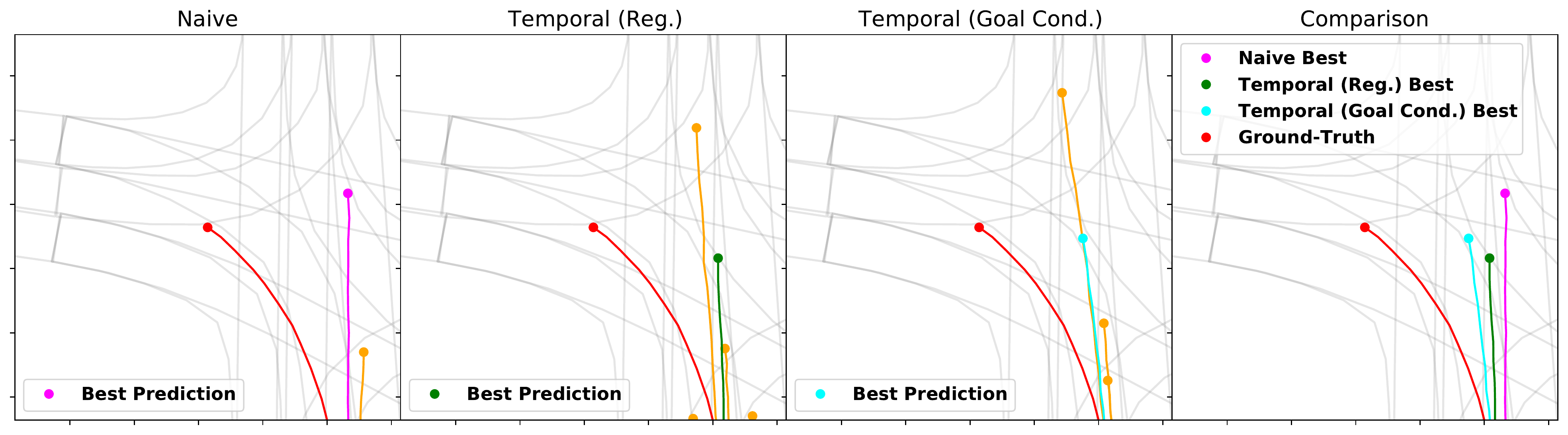}
    \end{subfigure} 
    
    \begin{subfigure}[h]{\textwidth}
      \includegraphics[width=1\linewidth]{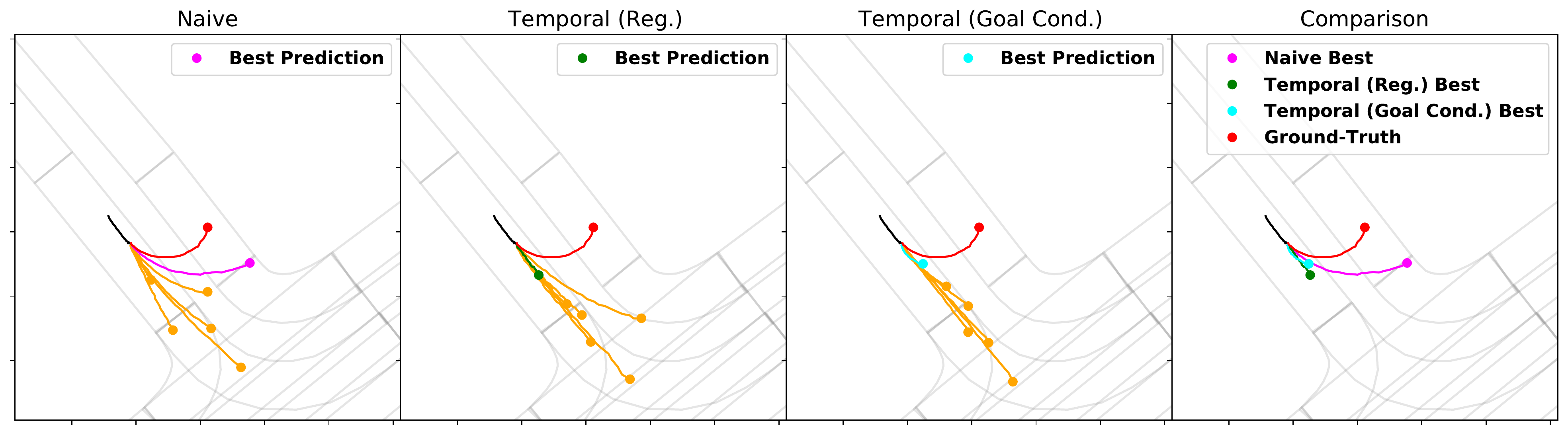}
    \end{subfigure}
    
\caption{\textbf{Fail cases caused by mode missing.} In some cases, our model could not catch the future intention of agent of interest vehicle such as left turn and U-turn.}
\label{fig:mode_missing_set}
\end{figure}

\begin{figure}[h!]
\centering
    \begin{subfigure}[h]{\textwidth}
      \includegraphics[width=1\linewidth]{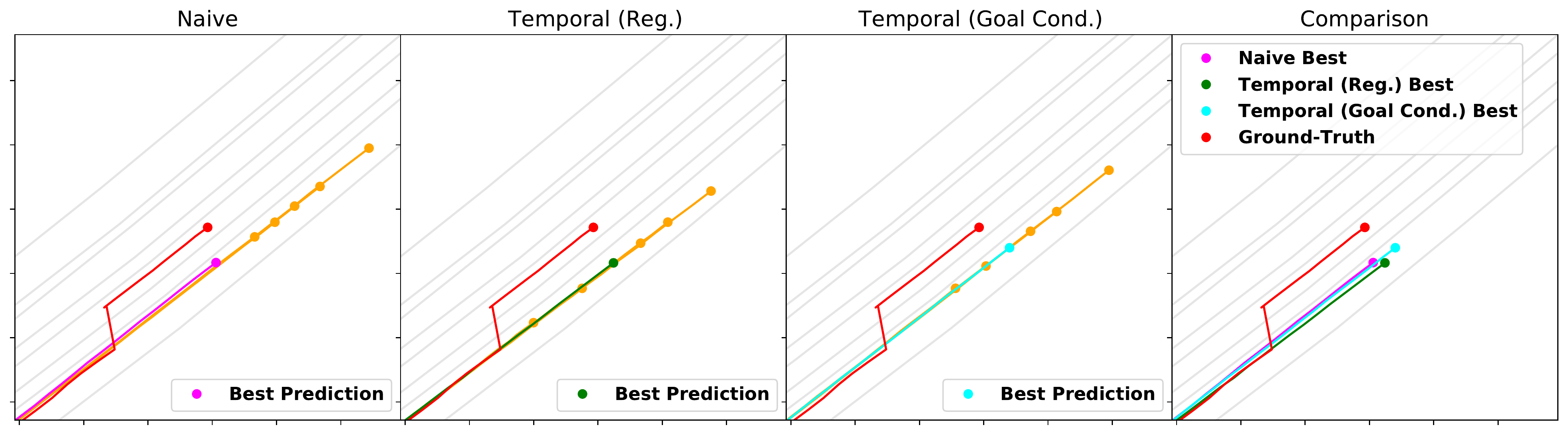}
    \end{subfigure} 

    \begin{subfigure}[h]{\textwidth}
      \includegraphics[width=1\linewidth]{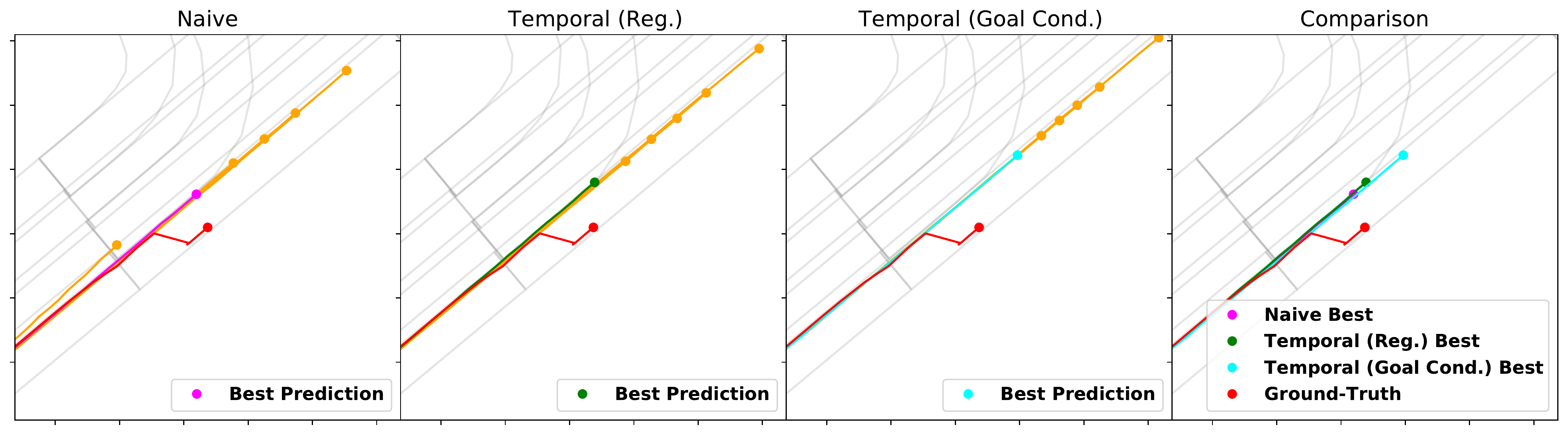}
    \end{subfigure}
    
\caption{\textbf{Fail cases caused by lane change.} In some cases, lane changes of agent of interest caused the fail. Since there is no sign about lane change on input sequences, lane changes remain hard to predict cases.}
\label{fig:lane_change_set}
\end{figure}

\begin{figure}[h!]
\centering
    \begin{subfigure}[h]{\textwidth}
      \includegraphics[width=1\linewidth]{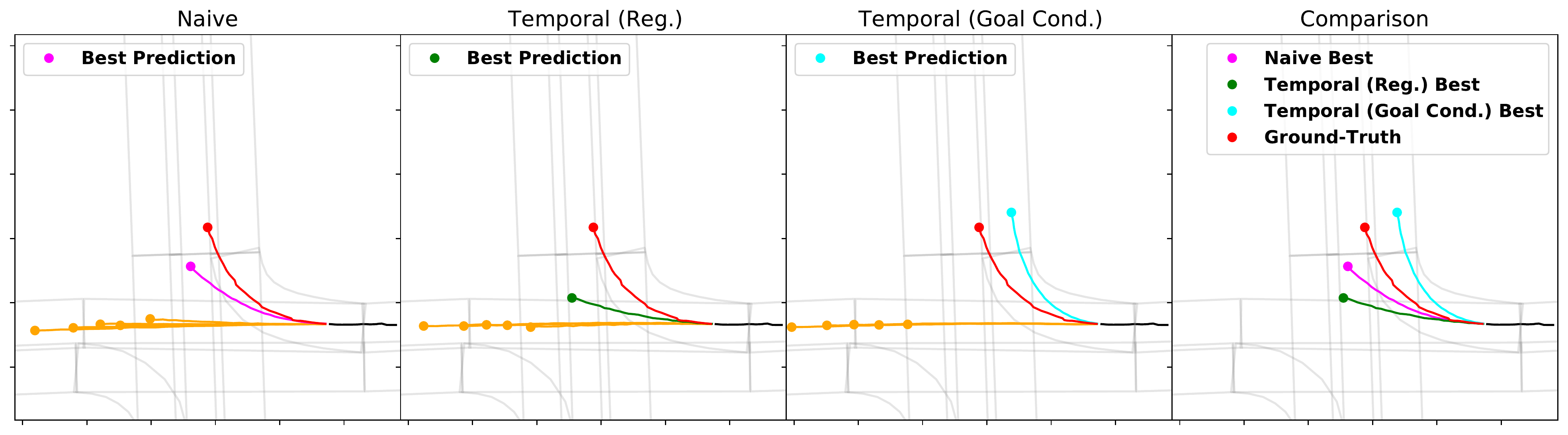}
    \end{subfigure} 

    \begin{subfigure}[h]{\textwidth}
      \includegraphics[width=1\linewidth]{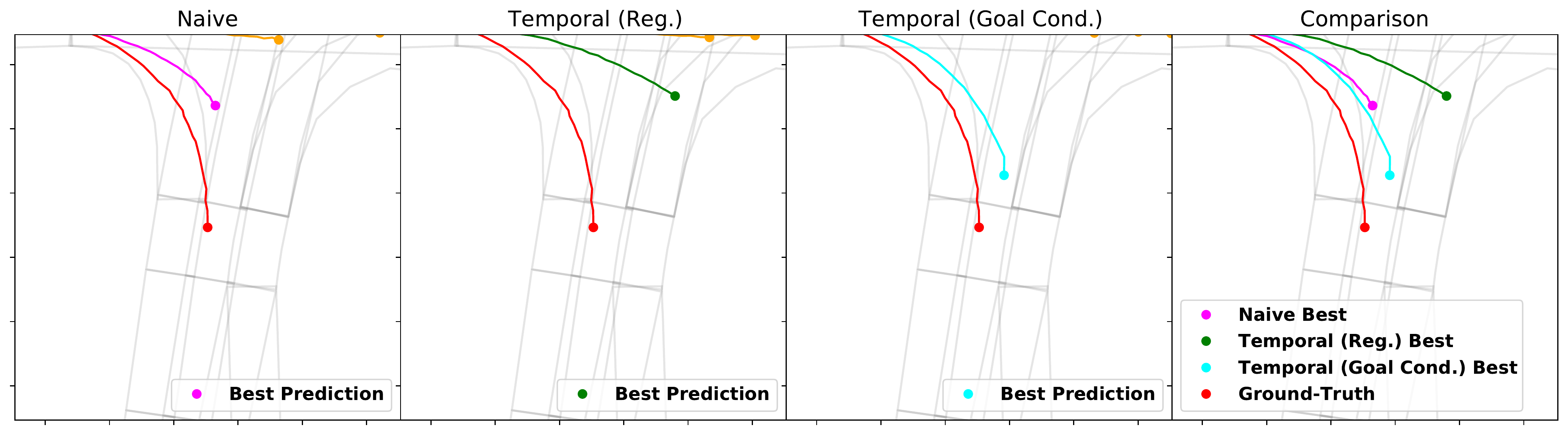}
    \end{subfigure}
    
\caption{\textbf{Fail cases caused by inaccurate prediction despite true mode prediction.} Although our model predicted the intention of agent of interest such as turns, it could not generate endpoints and trajectories close enough to ground truth.}
\label{fig:inaccurate_set}
\end{figure}

\begin{figure}[h!]
\centering
    \begin{subfigure}[h]{\textwidth}
      \includegraphics[width=1\linewidth]{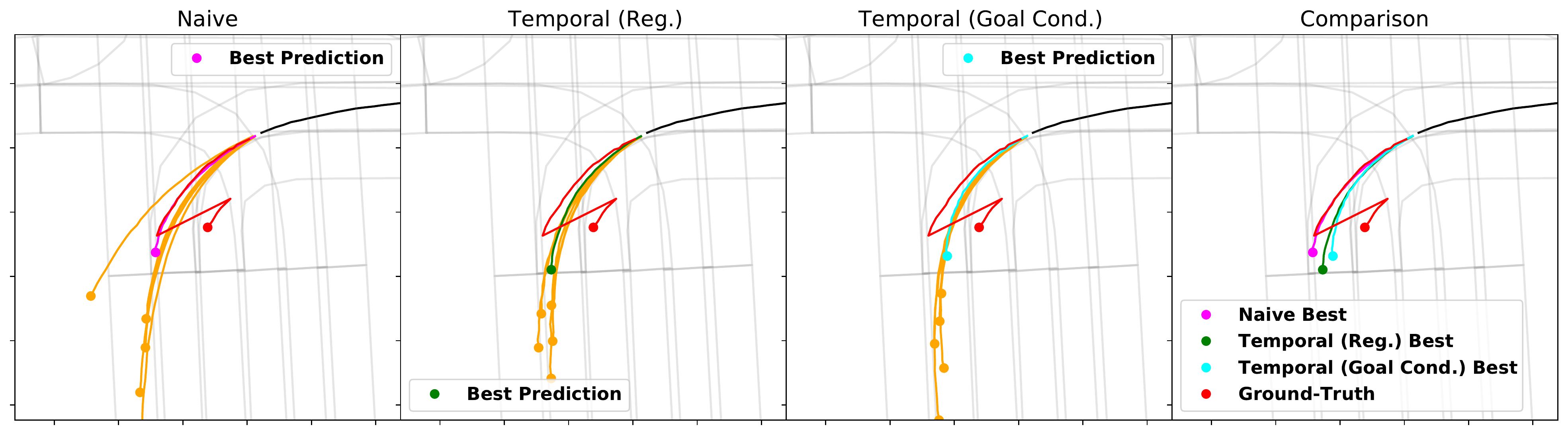}
    \end{subfigure} 

    \begin{subfigure}[h]{\textwidth}
      \includegraphics[width=1\linewidth]{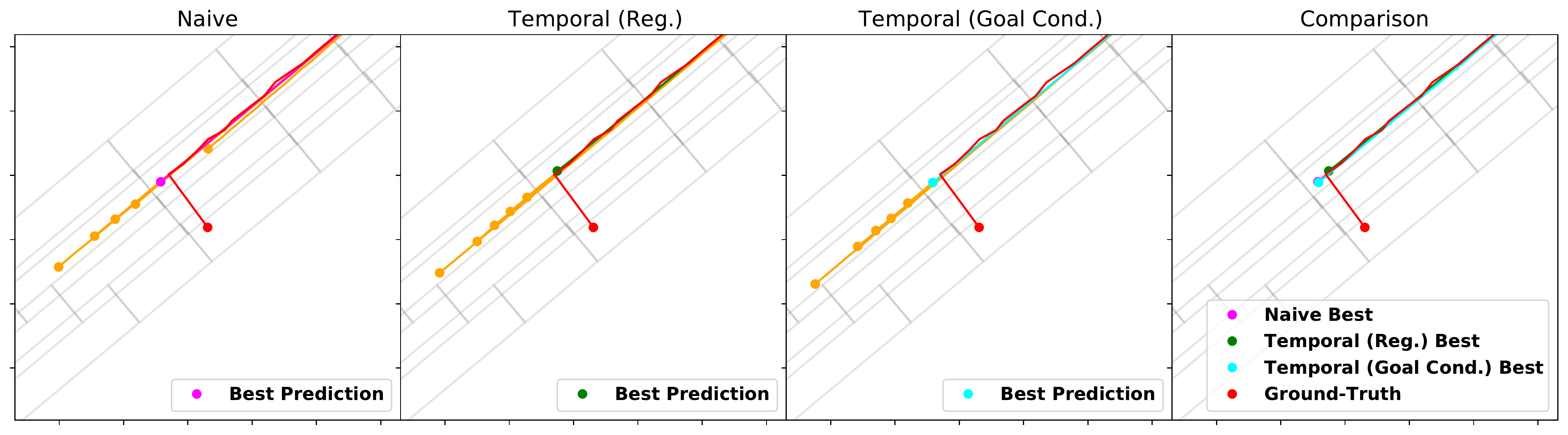}
    \end{subfigure}

    \begin{subfigure}[h]{\textwidth}
      \includegraphics[width=1\linewidth]{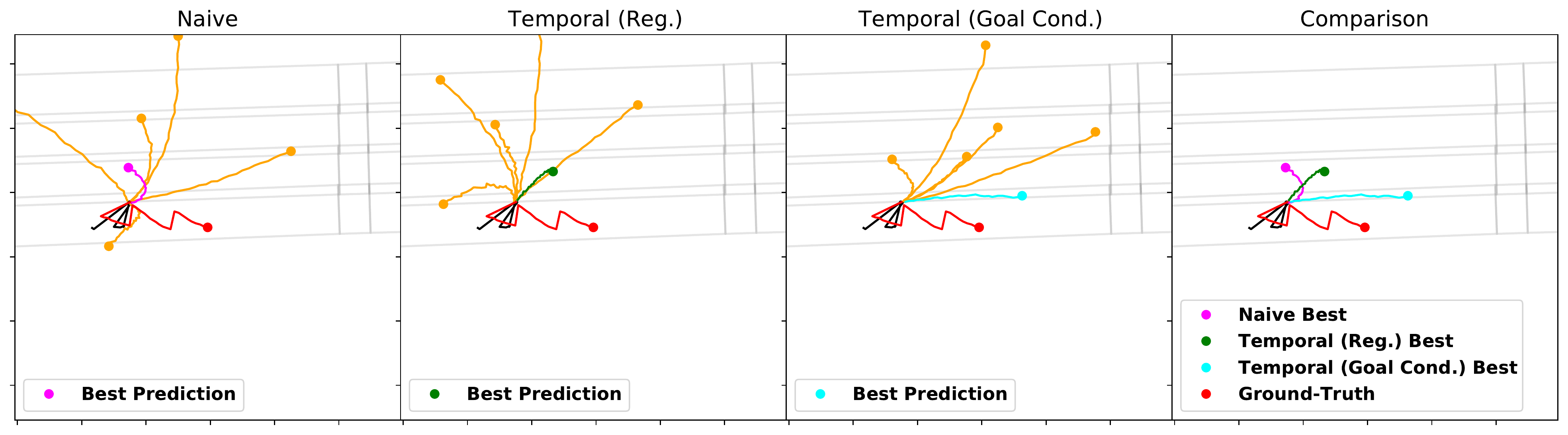}
    \end{subfigure} 

\caption{\textbf{Fail cases caused by data defects.} There are some faulty or uninformative input sequences as well as inconsistent ground truth sequences in the dataset resulting illogical trajectory predictions or high reported errors despite admissible predictions respectively.}
\label{fig:defect_set}
\end{figure}

In this section, we first provide some qualitative results of the same scene from regressive model without temporal encoding [Naive], regressive model with temporal encoding [Temporal (Reg.)] and goal conditioned model with temporal encoding [Temporal (Goal Cond.)] in \figref{fig:supp_comp_set1} and \figref{fig:supp_comp_set2}. Furthermore, we provide some fail cases with the reasons which are \textbf{mode missing} in \figref{fig:mode_missing_set}, \textbf{lane change} in \figref{fig:lane_change_set}, \textbf{inaccurate prediction} with true prediction of future mode in \figref{fig:inaccurate_set} and \textbf{data defects} in past input or future output sequence being also pointed out by other works ~\cite{Ye2021CVPR, Song2021CoRL} in \figref{fig:defect_set}

\end{document}